\documentclass{article}
\usepackage{ijcai20}

\usepackage{times}
\usepackage{soul}
\usepackage{url}
\usepackage[hidelinks]{hyperref}
\usepackage[utf8]{inputenc}
\usepackage[small]{caption}
\usepackage{graphicx}
\usepackage{amsmath}
\usepackage{amsthm}
\usepackage{booktabs}
\usepackage{algorithm}
\usepackage{algorithmic}
\usepackage{subfigure}
\usepackage{amsfonts}
\usepackage{multirow}

\title{Temporal-adaptive Hierarchical Reinforcement Learning}

\author{
Wen-Ji Zhou$^1$
\and
Yang Yu$^2$
\affiliations
$^1$Alibaba Group\\
$^2$National Key Laboratory for Novel Software Technology, Nanjing University
\emails
eric.zwj@alibaba-inc.com,
yuy@nju.edu.cn
}
\begin{document}

\maketitle

\begin{abstract}
	Hierarchical reinforcement learning (HRL) helps address large-scale and sparse reward issues in reinforcement learning. In HRL, the policy model has an inner representation structured in levels. With this structure, the reinforcement learning task is expected to be decomposed into corresponding levels with sub-tasks, and thus the learning can be more efficient. In HRL, although it is intuitive that a high-level policy only needs to make macro decisions in a low frequency, the exact frequency is hard to be simply determined. 
	Previous HRL approaches often employed a fixed-time skip strategy or learn a terminal condition without taking account of the context, which, however, not only requires manual adjustments but also sacrifices some decision granularity. 
	In this paper, we propose the \emph{temporal-adaptive hierarchical policy learning} (TEMPLE) structure, which uses a temporal gate to adaptively control the high-level policy decision frequency.
	We train the TEMPLE structure with PPO and test its performance in a range of environments including 2-D rooms, Mujoco tasks, and Atari games. The results show that the TEMPLE structure can lead to improved performance in these environments with a sequential adaptive high-level control.
\end{abstract}

\section{Introduction}
Reinforcement learning (RL) \cite{SuttonB98} aims at finding a policy with a maximum long-term reward from interactions with its environment. RL approaches commonly follow trial-and-error. The trial-and-error process often requires a large number of samples to find a good result. In complex tasks with, e.g., a high-dimensional state and action space, long horizon, and sparse reward, too many samples are required for traditional reinforcement learning methods to learn a good policy. 

It is noticeable that traditional RL approaches have many differences with the learning process of humans. One significant difference that may lead to the low-efficiency of the traditional RL approaches is the insufficient abstraction in the policy model. To address this issue, hierarchical reinforcement learning (HRL) studies RL with compositional models and has been developed over decades (see e.g., \cite{Barto2003}).
In general, HRL methods decompose a complex task into simpler sub-tasks, such that the whole task can be solved easier by utilizing the sub-task structure.

To manage the sub-task structure, some HRL methods use leveled structures in the policy model. In the two-level structure, the model is divided into a high-level part and a low-level part. While the low-level part is trained to output external actions, the high-level part is trained to output some internal actions to control the low-level part. In these methods, it is inevitable to consider the relationship between the two parts of the model. In some methods, when a low-level policy is chosen to run, it runs until termination, and only after that does the high-level policy choose the next low-level policy to run. For these methods, the terminal condition is crucial. Some methods use domain knowledge to design terminal conditions for each sub-policy manually \cite{SuttonPS99,ParrR97}.
However, it is time-consuming to design good terminal conditions for various tasks. Another way is to let the high-level policy make internal actions for controlling the low-level policy regularly. These approaches usually choose and switch the sub-policy at every certain time interval, and leave the time interval as a hyperparameter. Although this treatment is straight-forward, the interval is hard to be determined. Too large of an interval leads to a large decision granularity that may miss the critical controlling time. 
%For example, in an auto-driver system, late brakes can cause safety issues. 
Meanwhile, too small of an interval makes the learning of the high-level policy as frequent as the low-level policy, which degrades the HRL to be non-hierarchical. In recent years, some methods are proposed to learn a termination condition. But these methods often only consider the current state, while ignoring previous states and temporal information.

In this paper, we study the problem of how to choose a suitable control interval. We propose a temporal-adaptive hierarchical reinforcement learning (TEMPLE) approach that allows the policy to learn the timing for the high-level policy to control the low-level policy (also called sub-policy), without pre-definition. 
The main idea of TEMPLE is to introduce a temporal switch that can choose or mix between the current time and the previous time internal actions from the high-level policy, according to a switch signal. The switch signal is learned by the sub-policy, which can be viewed as a learning of the terminal condition. 

We implement TEMPLE with PPO \cite{ppo}. Then we validate TEMPLE on a range of different tasks, including grid-world room navigation environments, Mujoco \cite{TodorovET12} locomotion tasks and Atari games. 
These tasks are not easy to solve, since they have a long time horizon and relatively sparse reward.
Our method shows stable and improved performances than other methods.
%, including state-of-the-art hierarchical reinforcement learning approaches.

\section{Related Work}
Hierarchical reinforcement learning has attracted continuous attention in the past decades.
Previous works in hierarchical reinforcement learning seek to abstract actions or states manually.
For example, Options \cite{SuttonPS99} may be the most well-known formulation.
It abstracts actions to options.
An option is a sub-policy with a termination condition.
%which takes in environment observations and outputs actions until the termination condition is met.
There are also many other approaches besides Options.
The hierarchical automaton machine~\cite{ParrR97} compares automata with reinforcement learning. 
The MAXQ value function decomposition~\cite{Dietterich00} decomposes complex tasks into sub-goals.
These early studies~\cite{Hernandez2000,GhavamzadehM01} on hierarchical reinforcement learning approaches usually design the hierarchy manually, which costs a lot of manpower in every environment.
Many recent works focused on automatic hierarchical reinforcement learning \cite{VezhnevetsMOGVA16,DanielHPN16}. 
Option-Critic \cite{BaconHP17} learns options jointly with a policy-over-options \cite{SuttonPS99} in an end-to-end fashion by extending the policy gradient theorem to options.
SNN4HRL \cite{FlorensaDA17} learns a high-level policy and sub-policies, while sub-policies are trained by information-maximizing statistics.
FuN \cite{VezhnevetsOSHJS17} lets a high-level policy give goal directions to a low-level policy to help decision-making, and updates the two policy networks respectively. 
MLSH \cite{frans2017meta} learns on several similar environments to draw common sub-policies, and uses a high-level policy to choose them.
There are also many other hierarchical reinforcement learning works \cite{Ghazanfari17,OsaTS19,YangMJA18}. 
These hierarchical deep reinforcement learning methods can be divided into two general kinds. One kind learns several sub-policy networks and learns a high-level policy to choose them; an example of this kind is MLSH. The other kind uses a single neural network for all sub-policies; an example of this kind is FuN. Their high-level policy outputs high-level information to the sub-policy network, and the sub-policy network combines agent observation with these outputs to choose actions.

However, most of these automatic hierarchical reinforcement learning methods treat the timing of executing the high-level policy in simple ways. 
Most of them make high-level decisions at fixed steps.
This may make the high-level policy switch the sub-policy too often, which could degrade the HRL to be non-hierarchical, or too late,  which could harm the convergence of the policy and lead to inaccurate operations.
Although choosing the right timing is an important issue, there were few related works.
Notice that some option-based methods learn a termination condition to control the end of sub-policies. However, they do not take into account the sequence of the actions performed by the agent and only use the current state as the criterion to judge whether the sub-policy ends or not.
%In this paper, we focus on this issue and propose a novel temporal hierarchical reinforcement learning approach to learn switch timing automatically by utilizing a temporal structure.

\section{Temporal-adaptive Hierarchical Reinforcement Learning}

\subsection{Problem Statement}
%First, we formally define concepts in the hierarchical reinforcement learning problem.
Let $S$ and $A$ denote the state space and action space, respectively.
Let $P$ denote the transition function. 
$P(s'|s,a)$ means the probability of changing to state $s'$ when an agent executes action $a$ on state $s$.
Let $R$ denote the reward function, which is usually unknown to the agent.
$R(s,a)$ is the reward for executing action $a$ on state $s$.
Reinforcement learning aims at training an agent that can take actions to maximize discounted cumulative rewards. 

In hierarchical reinforcement learning problems, we discuss in this paper only the hierarchy of two levels for simplicity. Let $\pi_{high}(h|s;\phi)$ denote the high-level policy model, while $\phi$ is the parameter of the high-level policy.
The output of high-level policy $h$ is the internal action for controlling the sub-policy.
Let $\pi_{sub}(a|s,h;\theta)$ denote the sub-policy, while $\theta$ is the parameter of the sub-policy model and $h$ is from the high-level policy.
When running the policy, the high-level receives an observed state from the environment and outputs the internal action to the sub-policy; the sub-policy receives the state as well as the internal action, and then decides the output actions. In other words, the whole policy model can be presented as
\begin{equation}
\pi(a|s;\phi,\theta) = \pi_{sub}(a|s,h\sim \pi_{high}(s;\phi);\theta)
\end{equation}

It is easy to see that, when policy models are implemented as neural networks, the compounding of the high-level policy and the sub-policy forms a big neural network, which can be trained as a whole. However, this is just the same as the non-hierarchical method but does not make the learning problem simpler. A common idea is that the high-level policy does not make decisions as frequently as the sub-policy. Reducing the decision frequency of the high-level policy is a straightforward way to simplify the learning.

To control the decision frequency of the high-level policy, a switch can be introduced $h_t = \Omega(h_{t-1}, \hat h_t)$, where $\Omega$ determines whether the current internal action is the same as the previous time, or is the high-level policy output $\hat h_t \sim \pi_{high}(s_t;\phi)$. We can write the policy model as
\begin{equation}
\pi(a_t|s_t;\phi,\theta) = \pi_{sub}(a_t|s_t, h_t\sim \Omega(h_{t-1},\hat h_t);\theta)
\end{equation}

Previously, $\Omega$ was commonly designed as a fixed time switch. In the followings section, we introduce our temporal-adaptive design, inspired by the LSTM structure \cite{HochreiterS97}.

\begin{figure*}[h!]
	\begin{center}
		\center
		\includegraphics[width=0.65\textwidth]{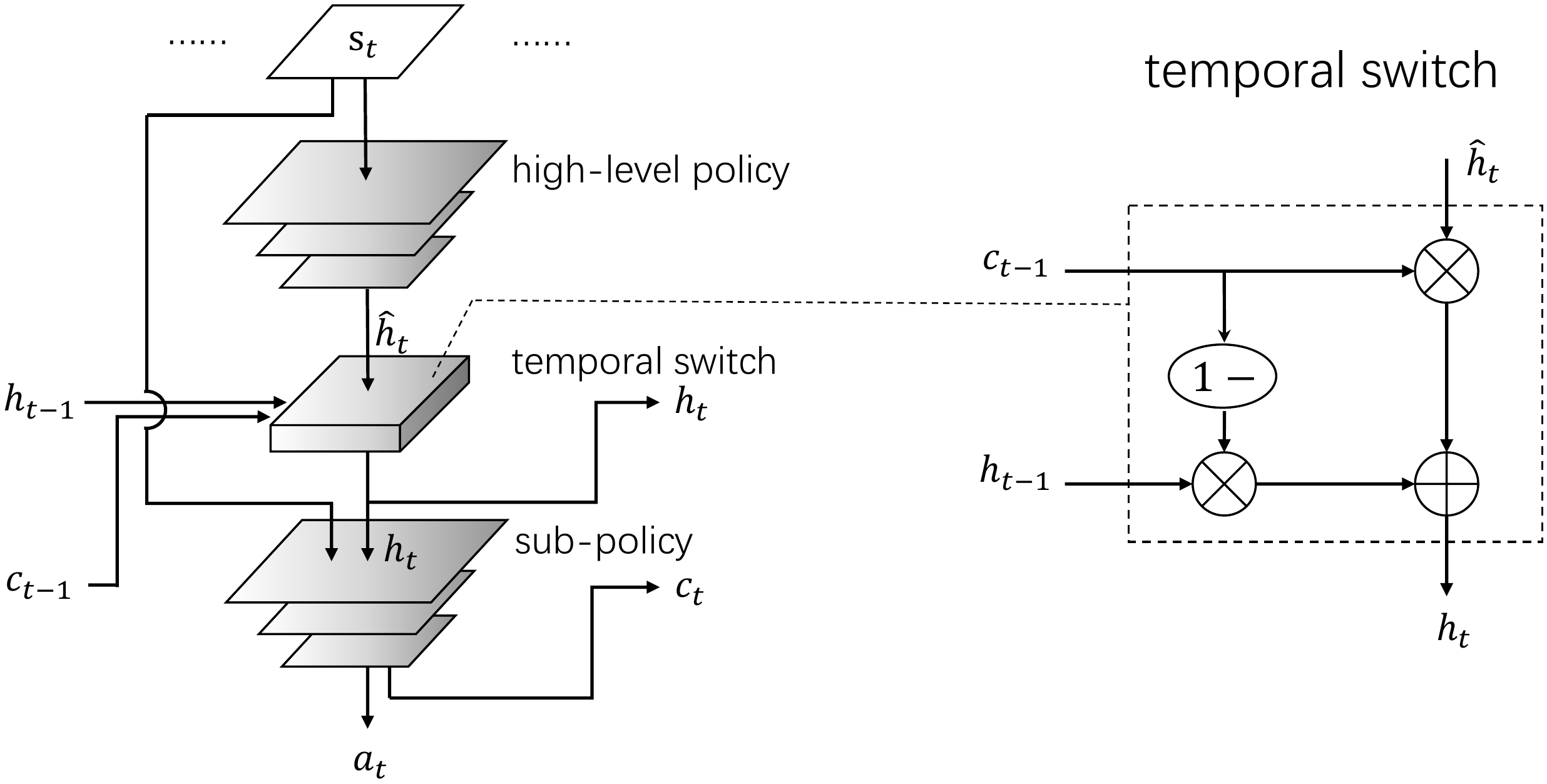}\\
		\caption{TEMPLE architecture. \emph{Left:} the hierarchical policy model structure with a temporal switch in between the high-level policy and the sub-policy; \emph{Right:} the inner structure of the temporal switch.}
		\label{fig:architecture}%\vspace{-1em}
	\end{center}
%\vspace{-0.8em}
\end{figure*}

\subsection{TEMPLE}
% Main idea
%In this section, we describe our formulation to solve the high-level decision frequency issue.
%We propose a temporal-adaptive hierarchical reinforcement learning (TEMPLE) approach.
The main idea of TEMPLE is to let the sub-policy determine if the internal action should be changed under the current context. The sub-policy outputs a signal to trigger the switch for the next time step. The policy structure is shown in Figure \ref{fig:architecture}. We also use a two-level HRL structure that consists of a high-level policy, a sub-policy, and a switch.

\subsubsection{High-level Policy}
%% detail process
As shown in Figure \ref{fig:architecture}, the high-level policy only receives current state $s_t$ as input. 
With its policy model parameters $\phi$, it produces the internal actions $\hat{h}$ according to $s_t$,
\begin{equation}
\hat{h}_t = \pi_{high}(s_t|\phi)
\label{eq_high_level_policy}.
\end{equation}
where $\hat{h}_t \in \mathbb{R}^d$ and $d$ is the dimension size of the internal action. 
A parameter to be determined is $d$, the dimensionality of $\hat{h}$. 
The larger the value of $d$, the richer the upper layer information contained in $\hat{h}$
We will investigate empirically the effect of this dimensionality.
%Some similar ideas have presented in other works.
%However, this method ignore the timing of switch sub-policies.
%As mentioned before, we propsed a temporal structure to solve this.

\subsubsection{Sub-policy}
We employ the single-model sub-policy with conditions, rather than separated-model sub-policies. Similar ideas are presented in \cite{FlorensaDA17,VezhnevetsOSHJS17}. At time $t$, the sub-policy network takes both the current state $s_t$ and the internal action $h_t$ as input, where the latter serves as the condition. Then it produces an action $a_t$ and a temporal gate signal $c_t$,
\begin{equation}
a_{t},c_{t} = \pi_{sub}(s_t,h_t|\theta),
\label{eq_sub_policy}
\end{equation}
where $c_t \in [0,1]$ and $h_t \in \mathbb{R}^d$.
%The internal action $h_t$ guides the sub-policy network to act as different sub-policies. 
The sub-policy directly outputs the low-level action $a_t$ that interacts with the environment.
At the same time, the temporal gate signal $c_t$ is passed to the next time step, guiding the switch to determine if it should choose to update its output, at the $t+1$ time step.

\subsubsection{Temporal Switch}
A common method to control the high-level decision frequency is to set a fixed time step to update the internal action $h$, which will sacrifice some decision granularity.
To solve this problem, we use a temporal gate to adaptively control the high-level decision frequency.
As shown in Figure \ref{fig:architecture}(right), the high-level policy output $\hat{h}$ is combined with the previous internal action $h_{t-1}$ to produce the new inner controlling action $h_t$, using the temporal switch signal $c_{t-1}$,
\begin{equation}
h_{t} = c_{t-1} * \hat{h}_t + (1 - c_{t-1}) * h_{t-1}
\label{eq_H},
\end{equation}
while $c_t \in [0,1]$. In this way, the current internal action $h_t$ comes from the mixing of the previous internal action $h_{t-1}$ and the current high-level policy output $\hat{h}_t$.
More previous internal action will remain if $c_{t-1}$ is close to 0; otherwise, the new high-level policy output $\hat{h}_t$ will take the place of the old one. In our implementation, the gate signal $c_t$ is the output of a sigmoid neuron, expecting that $c_t$ could be close to 0 or 1, acting as a switch. Nevertheless, we cannot be sure, unfortunately, that closing to 0 or 1 is better than a moderate value. Meanwhile, even for a moderate value of $c_t$, the high-level policy can receive a training signal from afterward steps, which can make the learning simpler and thus better.
%% high level analysis
%This sequential architecture let agent to tell himself when to %change sub-policy and how much should it change.
%By utilizing the switch signal from last time step, it can mix old hierarchical information and new hierarchical information in an adaptive and appropriate way.
%The sub-policy also predict a switch signal to next time step according to current hierarchical information and current actual situation.

Through such a temporal structure, the sub-policy can automatically feedback the current situation to the high-level policy.
The high-level policy can make decisions according to $h_{t-1}$, which contains temporal information from previous steps.
We can expand TEMPLE into sequences of different lengths.
Longer sequences make TEMPLE consider earlier information while shorter sequences make TEMPLE more concerned about the current situation.

\subsubsection{Training}

Note that the whole policy model is differentiable, we thus can optimize TEMPLE as a whole.
The hyperparameters, including sequence length $l$ for the unrolled training and internal action dimension $d$, should be set before training. 
To start an agent, we set initial $h_0=\{\frac{1}{d}\}^d$ and initial $c_0=0$.
Then we run the policy to collect trajectories. We can utilize various policy gradient-based deep reinforcement learning algorithms, e.g., \cite{SchulmanWDRK17,MnihBMGLHSK16,LillicrapHPHETS15}, to train the policy model.
We choose to use PPO in this paper.
When updating of the neural network model, like training LSTM, we unroll the policy model to $l$ sequential structures with shared network weights and apply a back-propagation method to update the network weights through time. 

%In this paper we use PPO as the basic reinforcement learning algorithm.

\section{Experiments}

In this section, we conduct several experiments to validate the performance of TEMPLE and compare our methods with state-of-the-art hierarchical reinforcement learning approaches.\footnote{The code of TEMPLE will be released after the review.}
In order to figure out how TEMPLE works, we also look into the learning process of TEMPLE and analyze the change in the internal action during the training.

In these experiments, we use PPO \cite{SchulmanWDRK17} as the basic reinforcement learning algorithm and apply it to our architecture.
We compare TEMPLE with the basic PPO algorithm using the same neural network structure as TEMPLE, serving as a baseline algorithm in our experiments.
It can be noticed that LSTM policy also uses temporal structure to capture sequential information in reinforcement learning tasks.
However, LSTM policy is commonly used at the state level to encode states, while TEMPLE uses a temporal gate in between the high-level and the low-level policies as a switch but not encodes the high-level policy output. In other words, TEMPLE does not apply an LSTM.
In order to distinguish TEMPLE from LSTM policy, we compare TEMPLE with state-based LSTM policy that encoding states using LSTM structure.
In order to verify the contribution of temporal switch, we also compared the TEMPLE-fix method.
In this method, the $\pi_{high}$ outputs the internal actions every fixed time. 
The $\pi_{sub}$ receive  internal actions and states and then produces the actions.
The $\pi_{high}$ and $\pi_{sub}$ are optimized by reinforcement learning separately. And the reward of $\pi_{high}$ is the total reward received from environments during the current internal actions action.

We also compare our method with the hierarchical reinforcement learning method A2OC \cite{HarbBKP18} and the meta-learning shared hierarchies (MLSH) method \cite{frans2017meta}.
Both of them are HRL methods that do not require manual design of sub-policies.
A2OC is an improvement of Option-Critic \cite{BaconHP17}. 
It also learns abstractions from tasks by combining Options \cite{SuttonPS99} with deep reinforcement learning.
Although A2OC learns a termination condition end-to-end, it only take current state into consideration and ignore information from previous high-level policy outputs.
Meanwhile, MLSH learns two-level abstractions.
The high-level policy and sub-policies are learned together on a set of similar environments, while the high-level policy is reinitialized at every certain episode in order to draw low-level common sub-policies for the given environment. Thus it is is a fix-schedule HRL approach.

In these experiments, if not specified, we set the number of skills to 4 for all these methods. The sequence length of TEMPLE is set to 4 by default. LSTM policy also set its sequence length as 4 for fair comparisons.
The MLSH makes a high-level decision every 10-time steps by the default setting of its code.
The results of all experiments are the average of repeated experiments.

\subsection{Experiments on FetchTheKey Environment}

\begin{figure}[ht!]
	\centering
	%\vspace{-0.8em}
	\includegraphics[width=0.9\linewidth]{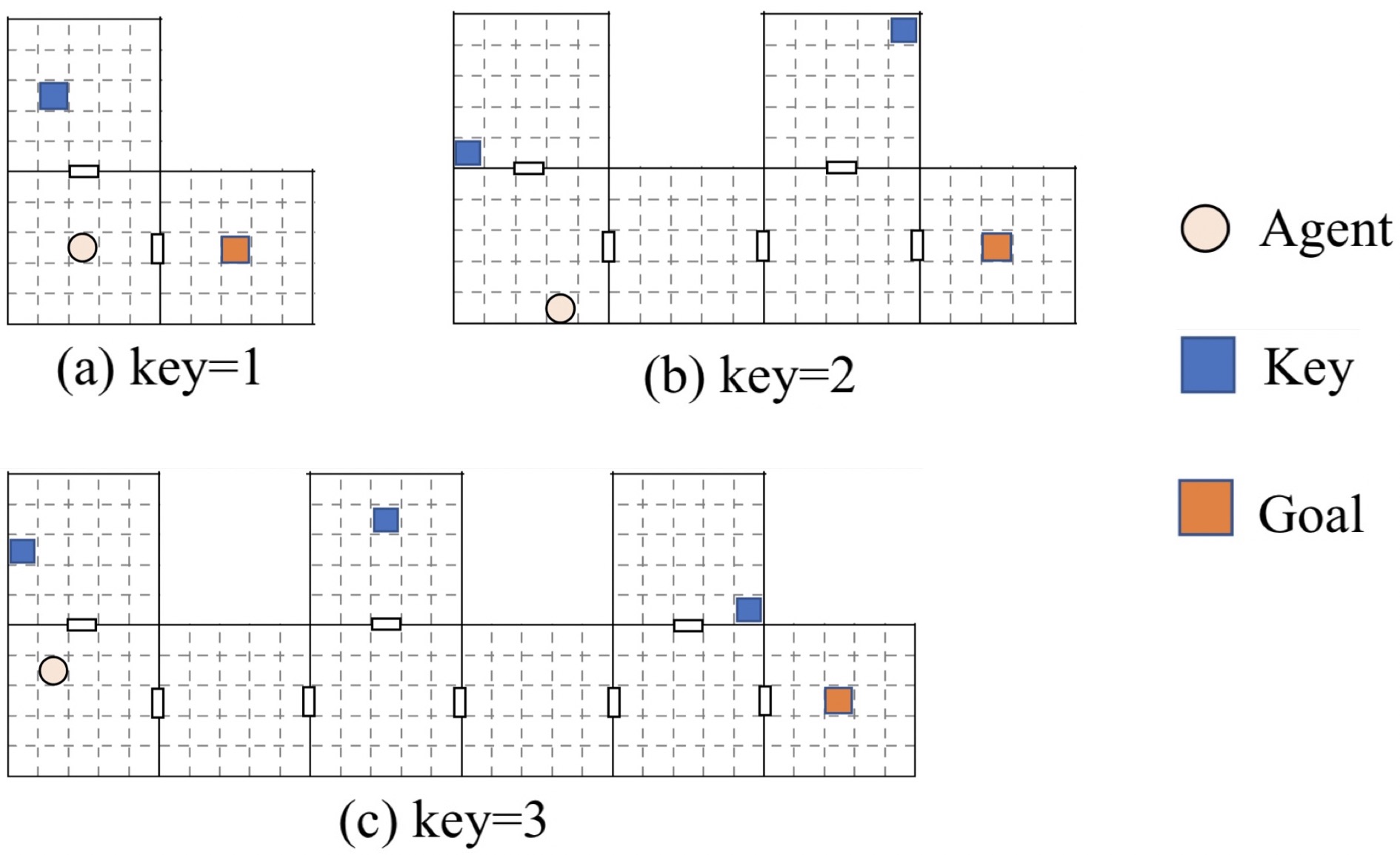}
	\caption{FetchTheKey environment}%\vspace{-0.5em}
	\label{fig:fetchthekey_env}
	%\vspace{-1em}
	\vspace{-0.5em}
\end{figure}

\begin{figure*}[ht!]
	\centering
	\includegraphics[width=4in]{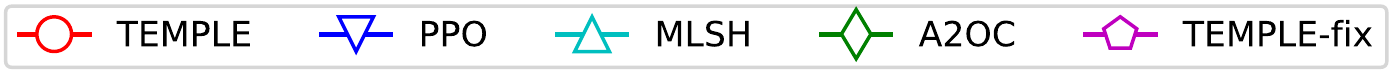}\\
	\subfigure[key=1 environment]{
		\includegraphics[width=0.23\textwidth, height=0.95in]{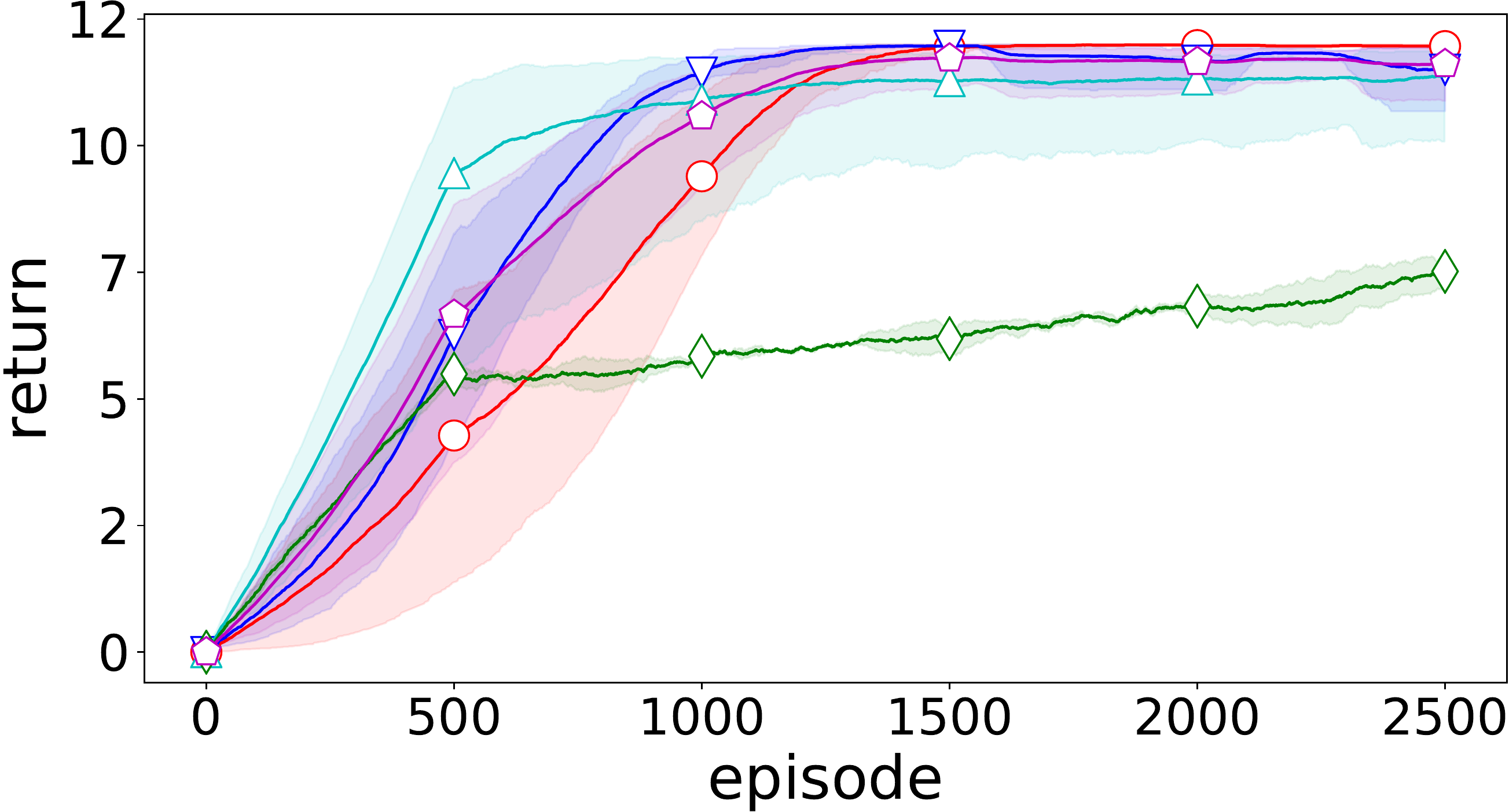}
	}
	\subfigure[key=2 environment]{
		\includegraphics[width=0.23\textwidth, height=0.95in]{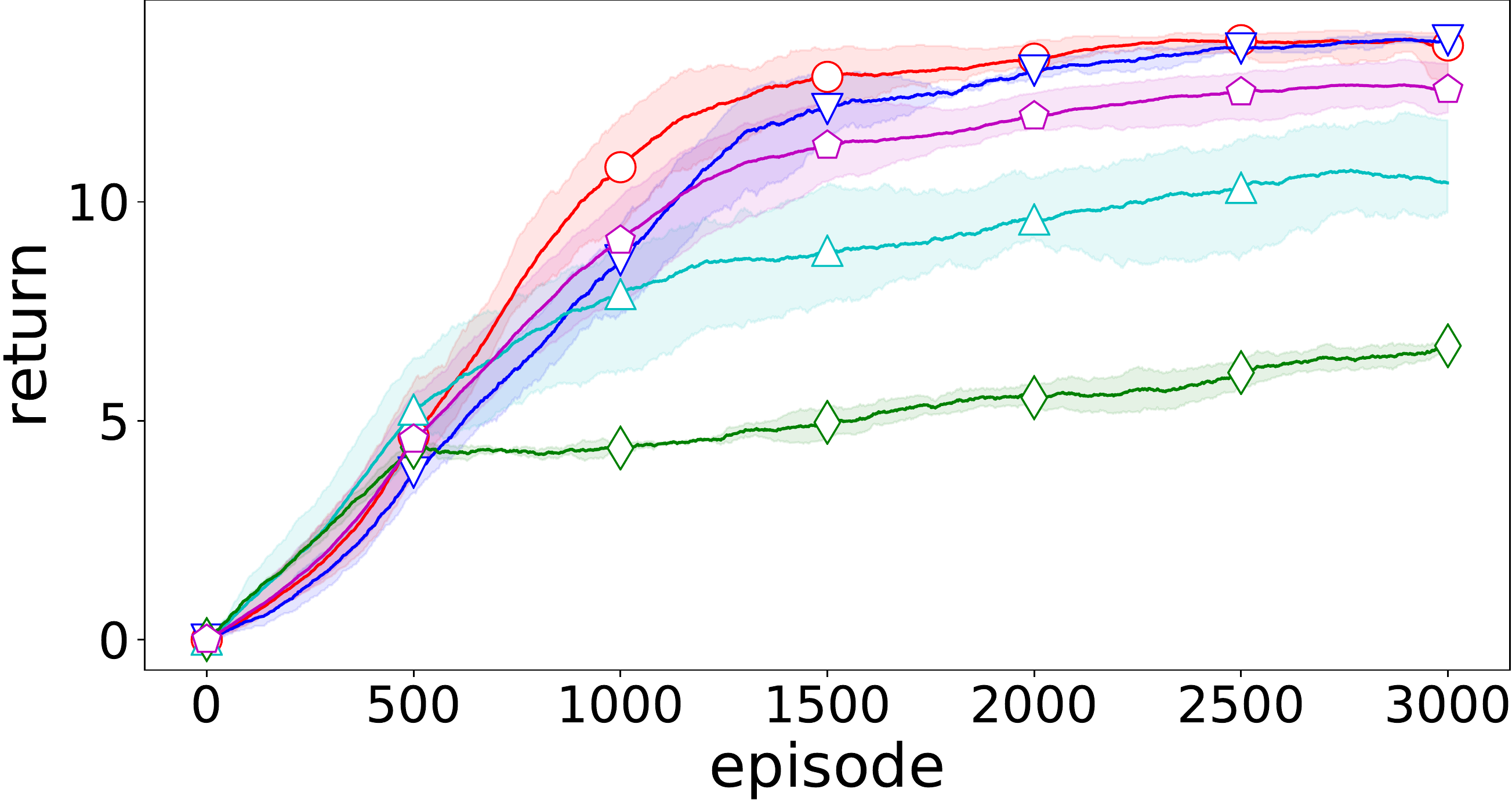}
	}
	\subfigure[key=3 environment]{
		\includegraphics[width=0.23\textwidth, height=0.95in]{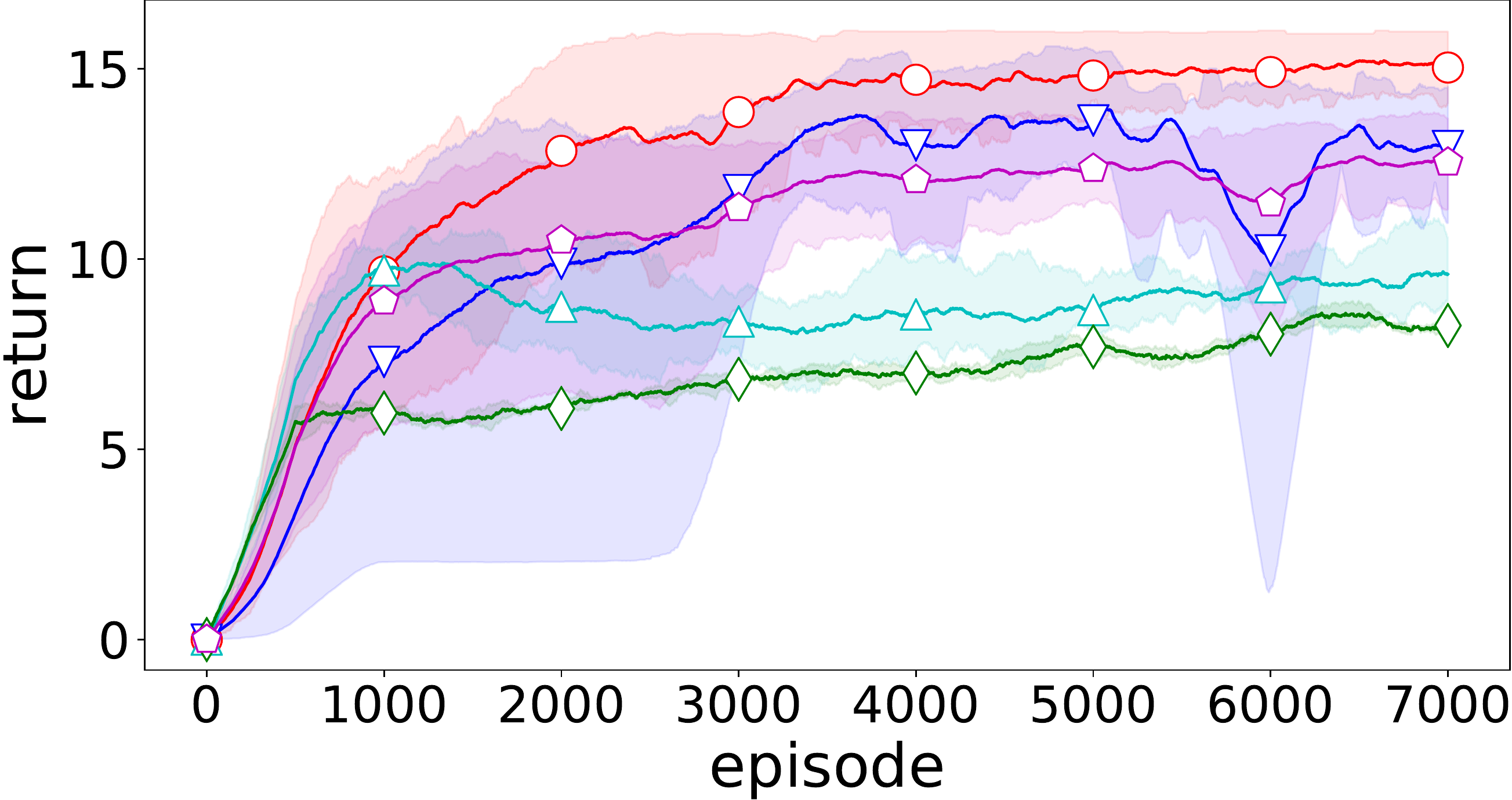}
	}
	\subfigure[key=4 environment]{
		\includegraphics[width=0.23\textwidth, height=0.95in]{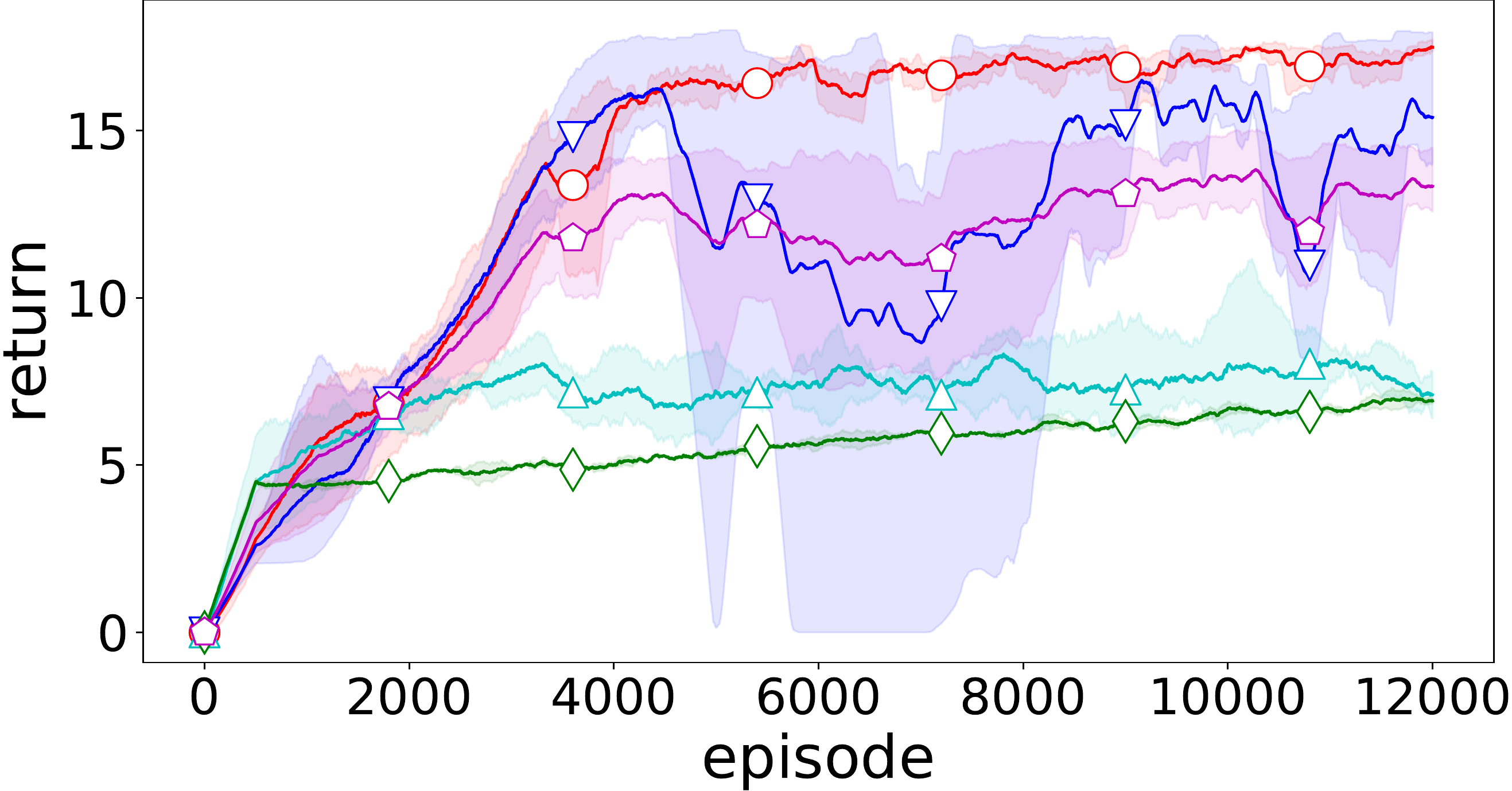}
	}%\vspace{-0.5em}
	\caption{Learning curves on FetchTheKey environment}%\vspace{-0.5em}
	\label{avg_ret_keyenv}
	\vspace{-0.5em}
\end{figure*}

First, we conduct experiments on the FetchTheKey environment. 
As shown in Figure \ref{fig:fetchthekey_env}, this is a grid world environment with many rooms.
The agent needs to fetch keys in different rooms and finally reach the goal position.
Rooms are connected by doors and the doors are closed at the beginning.
The agent can open the door only if it has the right key.
The agent can freely enter the room that contains the first key.
After that, the agent needs the key $i$ to enter the room that contains key $i+1$.
If the agent has all keys, it can enter the final room to reach the goal.
The positions of keys are randomly initialized in its room every time we reset the environment, while the position of the goal is fixed.
The agent can observe its position, the number of keys it picked up, and the grid information around it with a radius of 2. 
This means that the observation has 27 dimensions.
There are four dimensions of action: MOVE UP, MOVE DOWN, MOVE LEFT, and MOVE RIGHT.
The agent picks up keys automatically when it is at the position of a key.
The size of each room is $5\times5$.
The size of each door is $1\times1$, which is the same size as the agent.
Fetching keys gives a reward of 2 to the agent, and getting to the goal will give the agent a reward of 10.
The agent will not receive other reward signals.

This simple setting environment is, however, not easy to solve, because of narrow doors, a long horizon, and sparse reward.
Furthermore, additional keys to fetch means larger state spaces and longer horizons.
Thus, the difficulty increases as the number of keys increases.
%switch_env2
\begin{figure}[ht!]
	\centering
	\subfigure[key=2 environment]{
		\includegraphics[width=0.47\linewidth, height=1in]{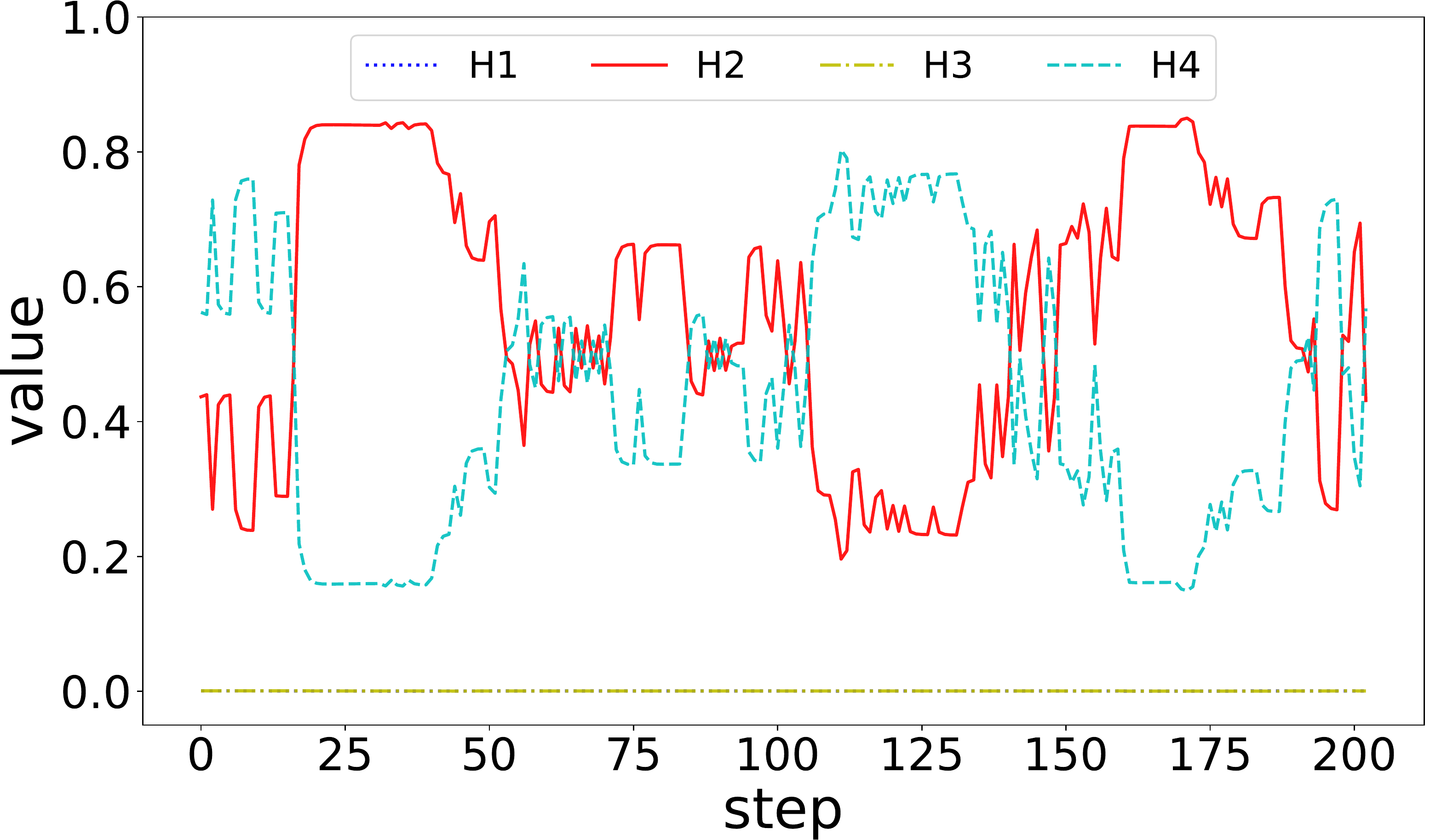}
		\includegraphics[width=0.47\linewidth, height=1in]{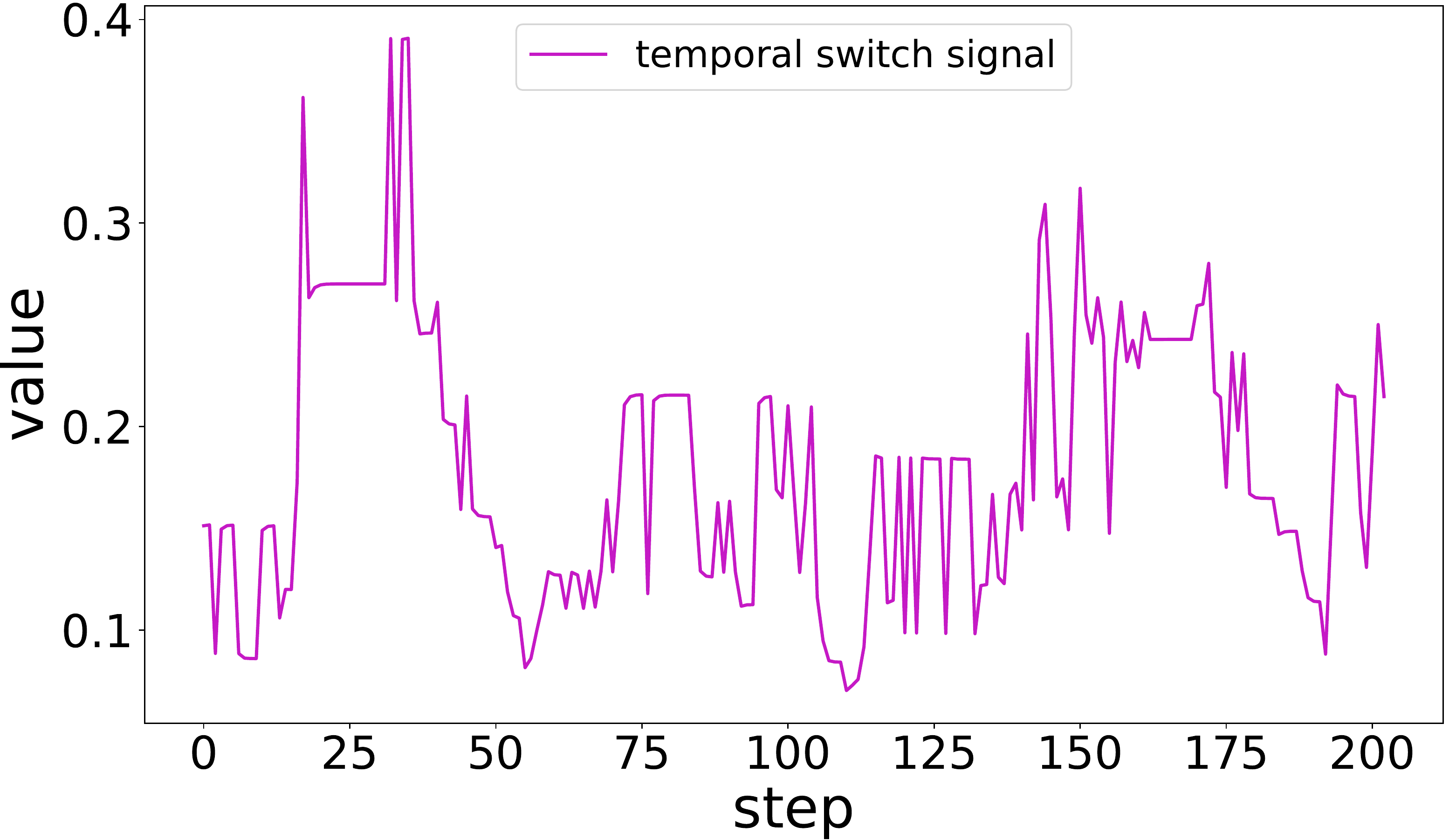}
	}\\
	\subfigure[key=3 environment]{
		\includegraphics[width=0.47\linewidth, height=1in]{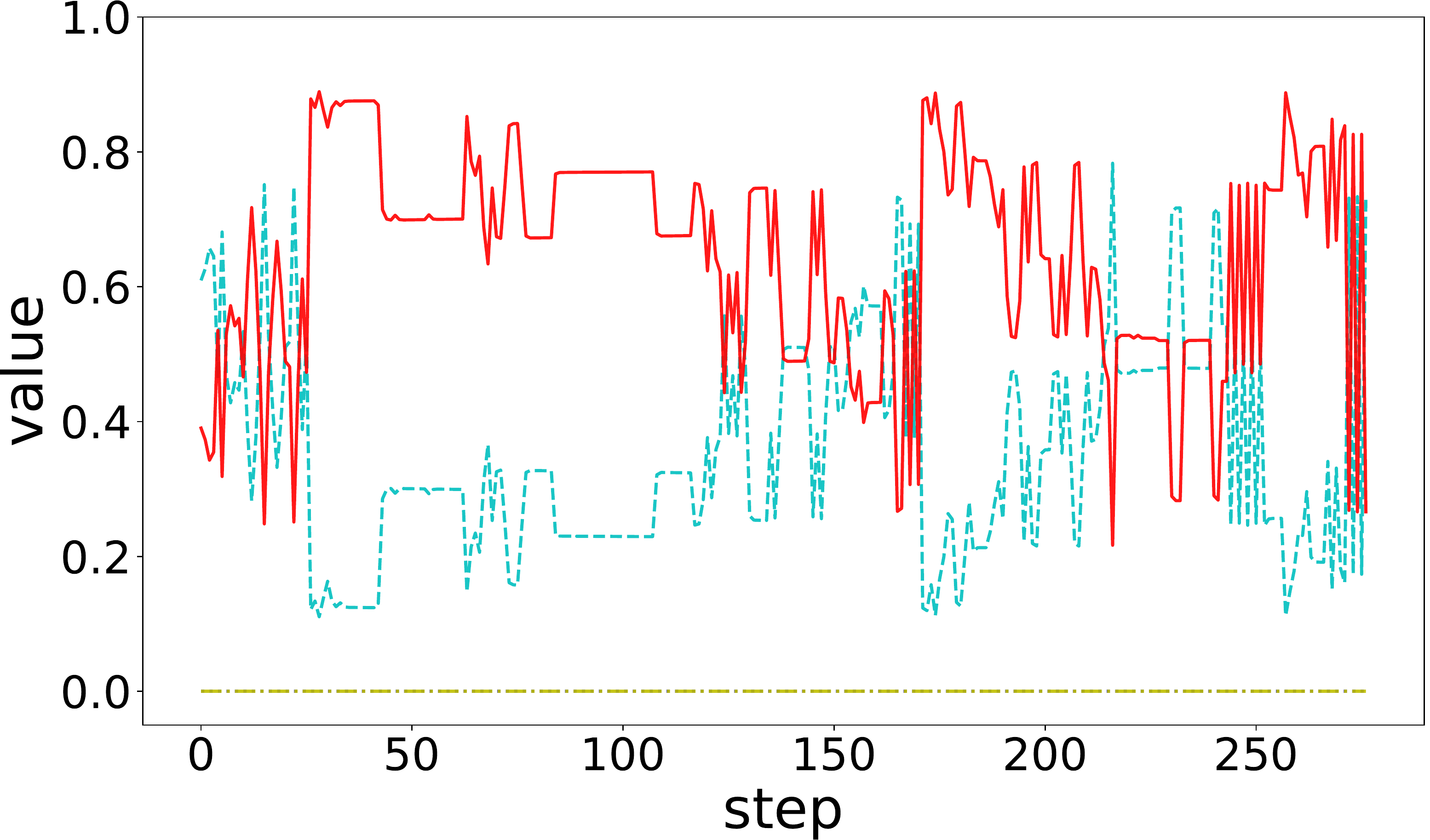}
		\includegraphics[width=0.47\linewidth, height=1in]{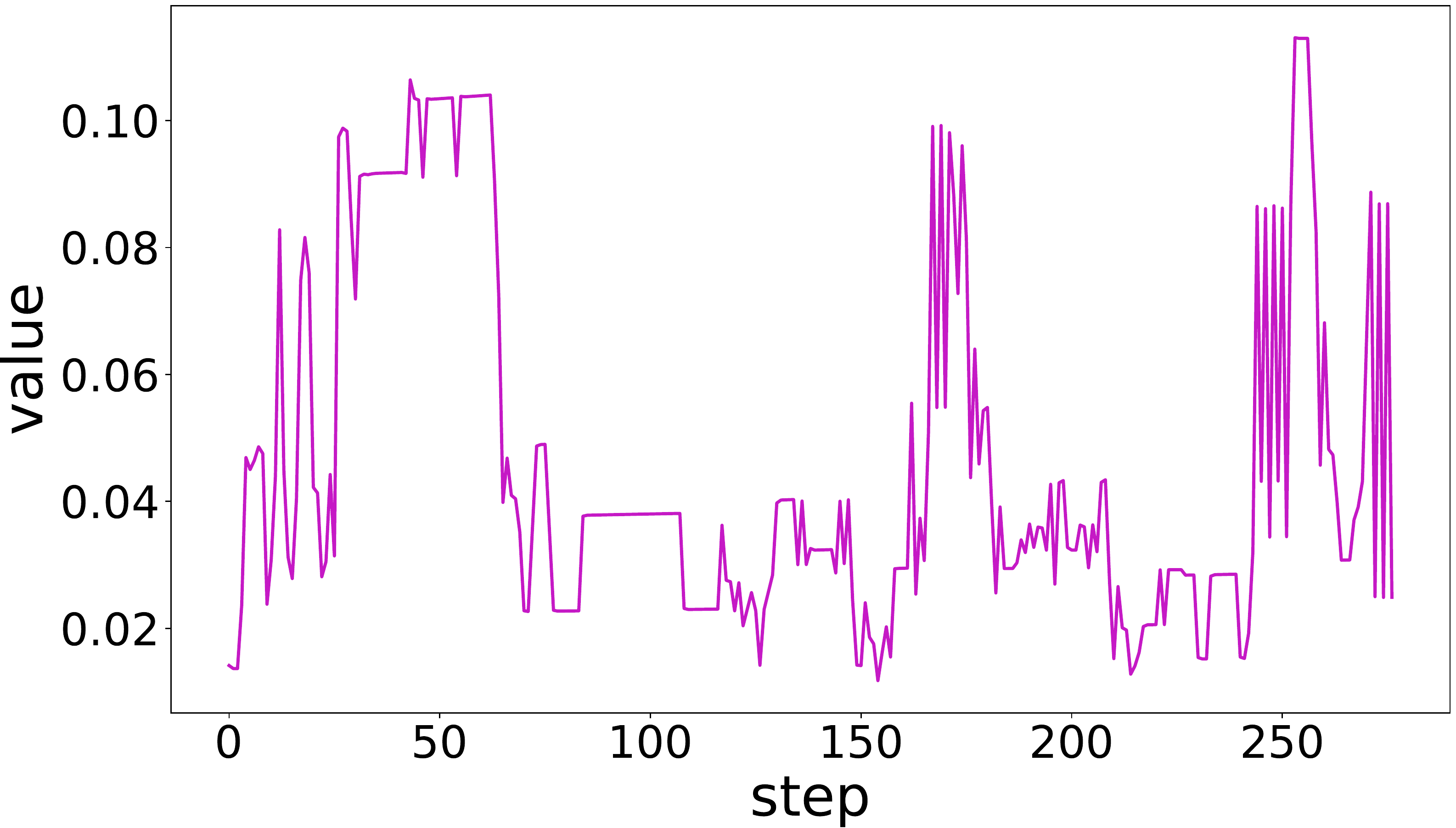}
	}\vspace{-0.5em}
	\caption{The changing of internal action and temporal switch in TEMPLE when playing on FetchTheKey environment.}
	\label{H_switch_analysis}
	\vspace{-0.8em}
\end{figure}

\subsubsection{Performance on FetchTheKey}
We first evaluate TEMPLE on different difficulties of the FetchTheKey environment: from the number of keys being 1 (key=1) to the number of keys being 4 (key=4).

The learning curves are shown in Fig \ref{avg_ret_keyenv}.
We notice that the performance of the basic PPO algorithm goes down with the increasing number of keys.
This indicates that the difficulty of environments actually grows with the number of keys that must be collected.
The basic PPO algorithm fails to learn a good policy and is hard to converge where there are three or four keys.
However, TEMPLE shows good performance on all of these environments.
The performance of MLSH is satisfactory when there is only one key. 
However, MLSH fails to have a good performance on the other environments.
This may be because MLSH can only switch the sub-policy at every certain step, leading to the loss of decision granularity.
%Moreover, its reinitialization of a high-level policy also slows the learning process.
LSTM policy is hard to learn good results, especially on harder environments. It means that a simple LSTM structure may not improve performance directly.
A2OC and FuN fail to learn good policies on all four environments. 
A2OC learns termination functions only according to current state and losses some temporal information. 
For example, when an agent is standing at a door, it is hard to judge whether to enter or exit the room based on the current state without considering the previous information.

%The superior performance of TEMPLE comes from its temporal and adaptive high-level decision-making structure.
%The flexible adaptive policy mechanism helps it better capture environmental details, and an appropriate switch timing leads to efficient utilization of sub-policies, which together lead to a better performance.

%On key=1 environment, all these methods have similar performance. 
%Because it is the easiest environment in these experiments.
%As the difficulty grows up, the differences between algorithms become obvious.
%When number of keys grows up to 2, although PPO and TEMPLE still have similar performance, TEMPLE shows a faster learning process than PPO.
%However, the return of MLSH grows up through the training process, but it starts to lag behind that other approaches.
%Its learning process is slow due to the reinitialization of high-level policy.
%On key=3 environment and key=4 environment, the PPO fails to learn a good policy and is hard to converge.
%It indicates that basic PPO algorithm can't handle such difficult environments, and hierarchical reinforcement approaches can have a better performance.
%However, MLSH converges to a bad policy soon.
%The fixed frequency of high-level policy decision making lets MLSH sacrifice some decision granularity.
%And A2OC also fails to learn good results.
%However, TEMPLE still has a good performance.
%It learns temporal hierarchical information to help it solve long horizon and sparse reward problems in difficult environments.

\subsubsection{Analysis of Internal Action}
We conduct several experiments to analyze the learning process of TEMPLE and the interpretability of its internal action by looking into the change in the internal action and the temporal gate signal during the test phase.

We set the internal action dimension to 4 and set the sequence length to 4.
We train and test TEMPLE on the key=2 and key=3 environments.
During the testing phase, we record the internal action $h$ and temporal gate signal $c$.

%\begin{figure*}[ht]
%\centering
%\subfigure[key=2 environment]{
%    \includegraphics[width=0.235\textwidth]{pic/H_env2.pdf}
%    \includegraphics[width=0.235\textwidth]{pic/switch_env2.pdf}
%    }
%\subfigure[key=3 environment]{
%    \includegraphics[width=0.235\textwidth]{pic/H_env3.pdf}
%    \includegraphics[width=0.235\textwidth]{pic/switch_env3.pdf}
%    }\vspace{-0.5em}
%    \caption{Internal action and switch signal in FetchTheKey environment.}
%    \label{H_switch_analysis}
%\end{figure*}

The change in the internal action and the temporal switch signal in a single test episode are shown as Figure \ref{H_switch_analysis}. First, we can observe that only two dimensions are active during the test process, the other two dimensions stay near zero, in both of the environments. This implies that TEMPLE can automatically distill redundant internal actions, so redundant internal actions do not have a negative influence on performance.
It can be observed that the internal action shows several different patterns during prediction.
The internal action keeps a pattern for some steps and changes to the next pattern, which implies that TEMPLE switches from one sub-policy to another.
Additionally, the changes in the temporal gate signal also match the change in the internal action.
%\begin{wrapfigure}{r}{0.45\textwidth}
\begin{figure}[ht]
	\centering
	%\vspace{-0.8em}
	\includegraphics[width=0.95\linewidth]{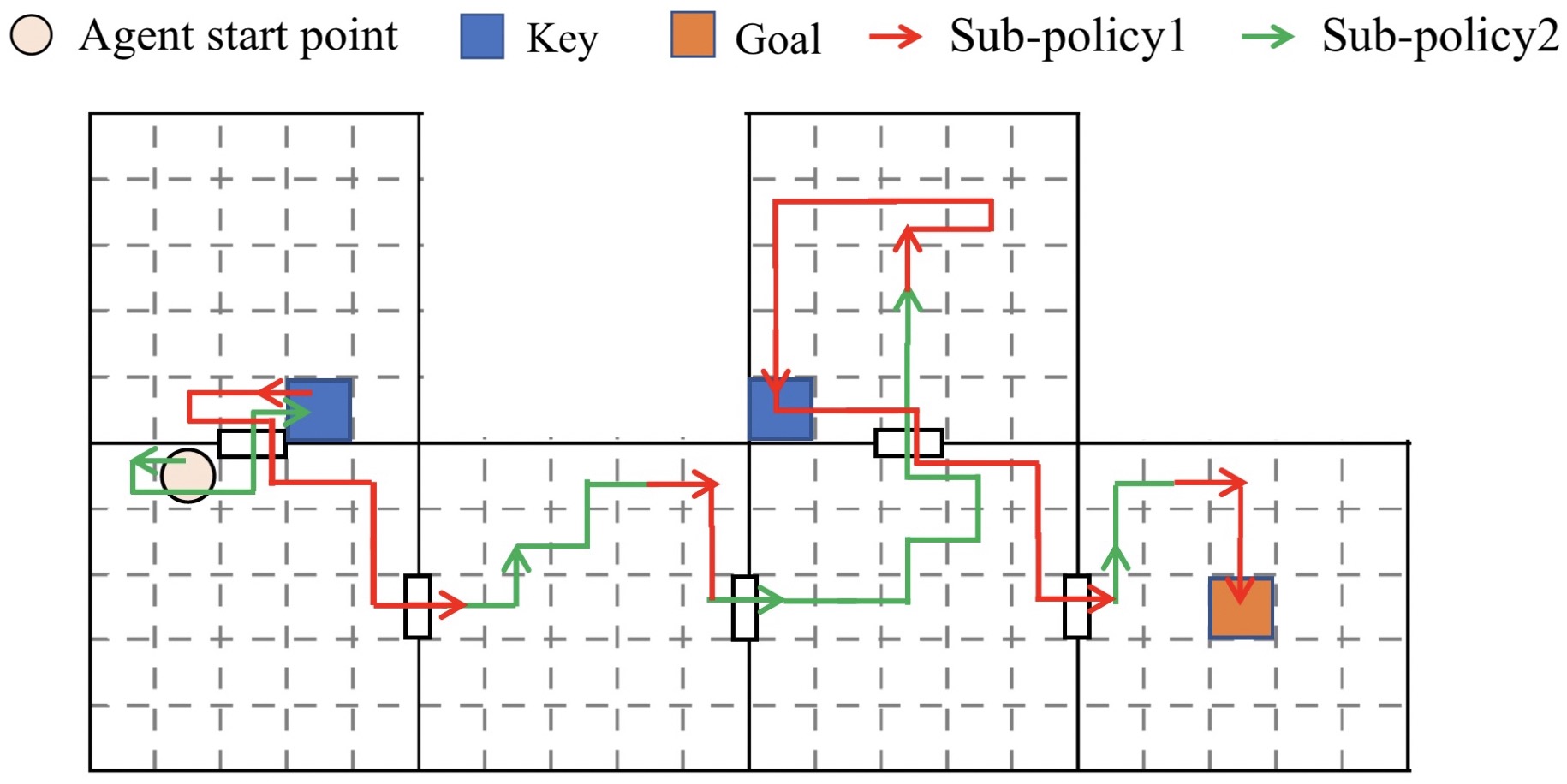}
	\caption{Trajectories of different sub-policies on the key=2 environment}%\vspace{-0.5em}
	\label{fig:sub-policy_traj}
	%\vspace{-1em}
	\vspace{-0.8em}
\end{figure}
%\end{wrapfigure}
Every time the temporal gate signal increases, the internal action undergoes significant changes.
This shows that the temporal gate signal influences the change in the internal action, and therefore the high-level policy decision frequency can be adaptively controlled by the temporal gate signal.
Besides, the frequency of internal action changes is not constant, which indicates that TEMPLE can dynamically and adaptively control the high-level policy decision frequency.

In order to intuitively understand the pattern of the sub-policy, we plot one trajectory in Figure \ref{fig:sub-policy_traj}. We color every step of the trajectory according to the value of the internal action dimensions: when H2 $>$ H4, the step is plotted in red, and otherwise the step is colored in green.
The moving direction is indicated by the arrows in the trajectory. It can be observed that the red segments and the green segments show different intentions of the agent. By the red segments, the agent tries to leave the current room and move left down to the next room. While by the green segments, the agent tends to move to the right, enter a room and search for keys. This result support that TEMPLE can learn meaningful hierarchical internal action with effective abstraction.

\subsubsection{Analysis of the Internal Action Dimension and the Sequence Length}
Notice that in the previous subsection we had four dimensions for the internal action but only two dimensions are active, i.e., non-zero.
This indicates that we may have some redundant dimensions.
However, it is hard to know the exact number of sub-policies before applying HRL to a new task.

\begin{figure}[ht]
	\centering
	%\subfigure[]{
		\includegraphics[width=0.495\linewidth, height=1.1in]{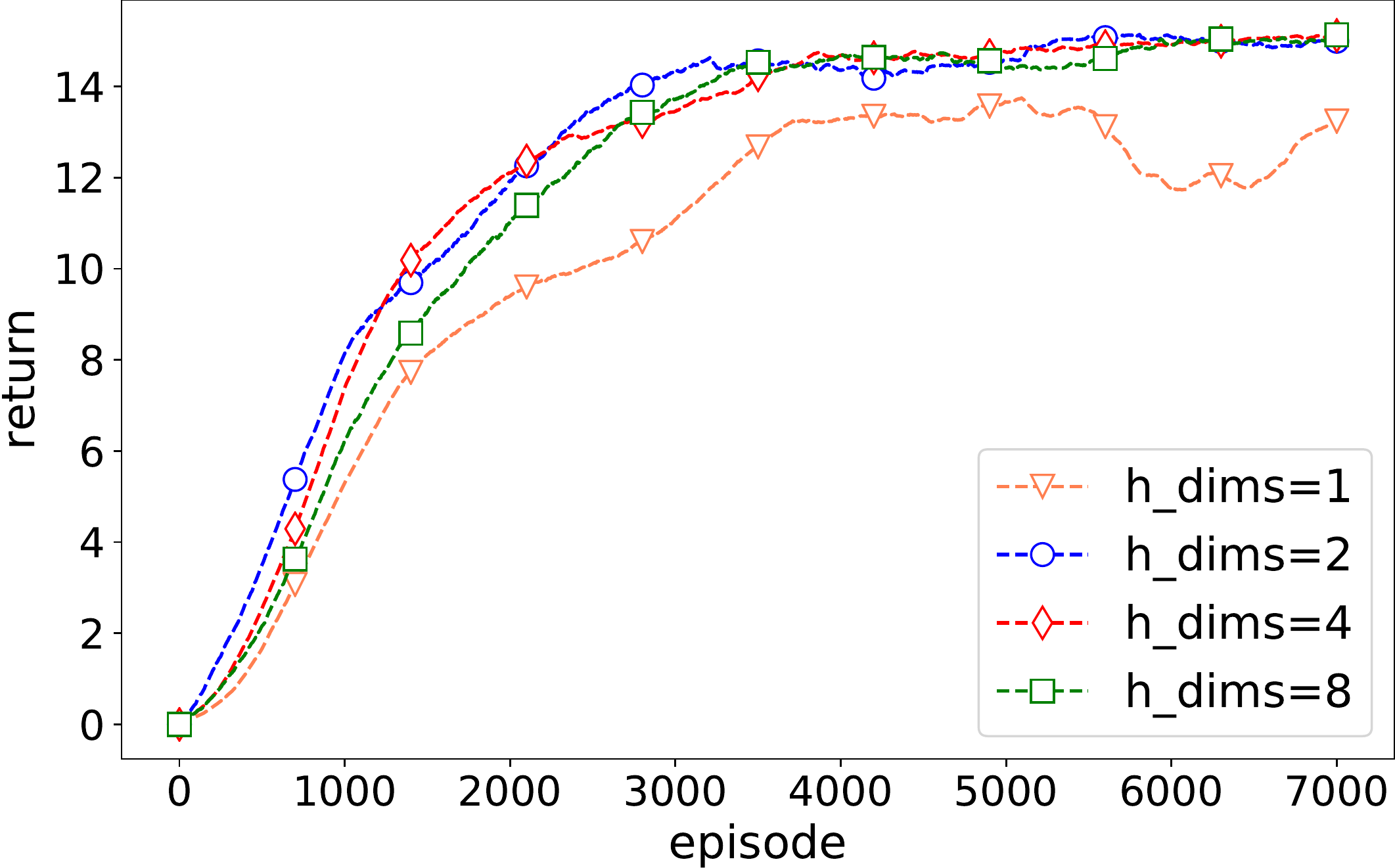}
	%}
%	\subfigure[]{
		\includegraphics[width=0.495\linewidth, height=1.1in]{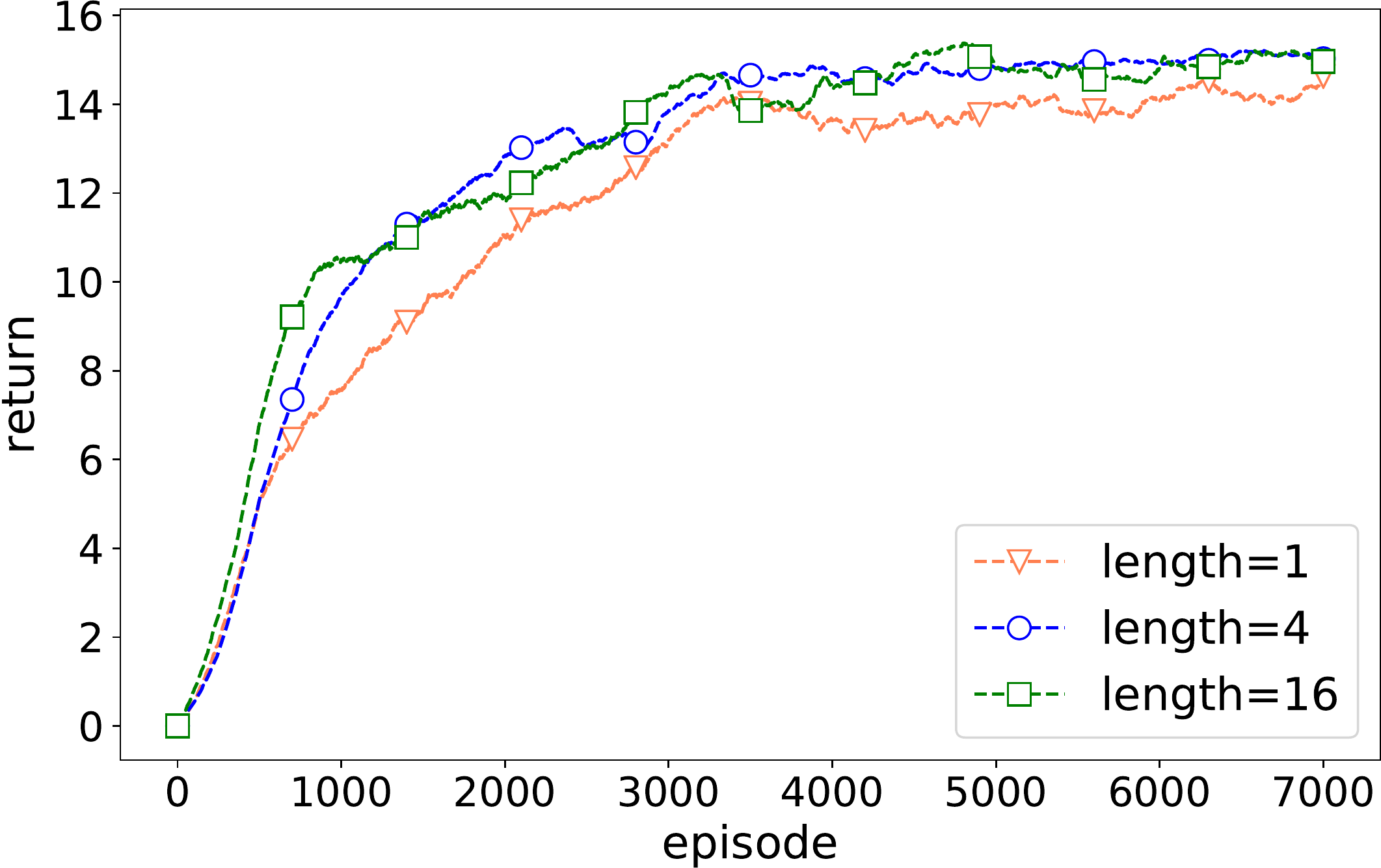}
%	}
	\caption{Learning curves on key=3 environment with different dimension size of internal action (left) and sequence length (right).}
	\label{fig:comp_skill_seq}
	\vspace{-0.5em}
\end{figure}

Another important hyper-parameter of TEMPLE is the sequence length for the unrolled training.
The sequence length of TEMPLE indicates the correlation between the current internal action and the previous information.
A longer sequence length means that the previous state and the internal action play a more important role in current high-level decision-making.
Therefore, we conduct experiments to analyze how the number of sub-policies and the length of TEMPLE would influence the training process of TEMPLE.  
%The details of settings and results of experiments are shown in Figure \ref{fig:comp_skill_seq}. 
We have several experiments on key=3 environment with different internal action dimensions and sequence lengths.

As shown in Figure \ref{fig:comp_skill_seq}, the results indicate that TEMPLE is not sensitive to redundant sub-policies and is robust. It can automatically distill the sub-policy models and reduce the redundancy. The results also show that the temporal structure is essential for high-level decision-making and that TEMPLE is robust to redundant sequence lengths.

\subsection{Experiments on Mujoco Environments}
\begin{figure}[ht]
	\centering
	%\vspace{-0.5em}
	\includegraphics[width=0.82\linewidth, height=1.6in]{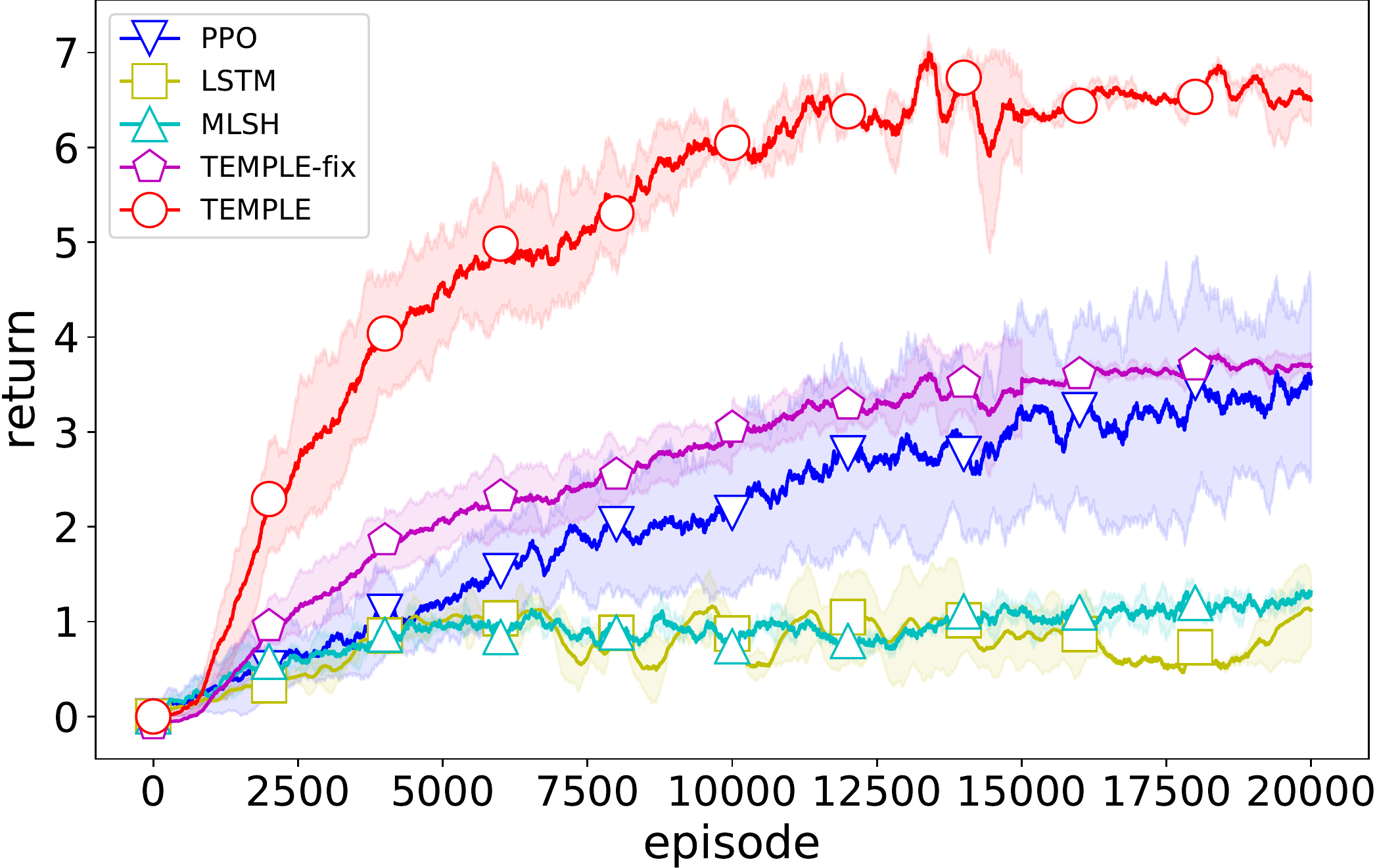}
	\caption{Learning curves on SwimmerGather environment}
	\label{fig:SG_result}
	\vspace{-0.8em}
\end{figure}

%\begin{wrapfigure}{r}{0.45\textwidth}
%\end{wrapfigure}
We also validate our approach on robot control tasks.
%The SwimmerGather environment is a hard robot control task on the Mujoco physics engine \cite{TodorovET12}.
The SwimmerGather environment is a difficult robot control task on the MuJoCophysics engine~\cite{TodorovET12}. It is also used to evaluate the performance of HRL methods in some previous works~\cite{FlorensaDA17}.
The agent is a snake and moves by twisting its body.
The states are observed by a sensor and the actions are the angles of the snake's body joints.
The agent gets a reward of 1 for collecting green balls and a reward of -1 for collecting the red ones.
There are no other reward signals. In this experiment, we set the SwimmerGather environment with 18 green balls and 5 red balls.
Since the official codes of A2OC only implement the discrete action space version, we here compare our proposed method with the basic PPO algorithm, LSTM policy, and MLSH method.
We also set the number of sub-policies to four for MLSH and TEMPLE.
All these methods train on the SwimmerGather environment for 20000 episodes, and the agent moves 3000 steps in a single episode.
The results are shown in Fig \ref{fig:SG_result}.

The basic PPO algorithm shows a slow learning process, which indicates that this environment is not easy for it.
The MLSH, LSTM and TEMPLE-fix policy also show slow learning processes and fail to learn good policies.
%It may be because MLSH is sensitive to the frequency of high-level decision making.
Our method shows superior performance over the other methods.
Due to the temporal-adaptive approach, the high-level policy knows when to change sub-policies and gives a more accurate internal action to the sub-policy, which finally results in a good performance.
This will lead to an efficient hierarchical reinforcement learning, which is especially suitable for solving some tough environments.

\subsection{Experiments on Atari Games}
%We also validate TEMPLE on Atari games.
%The Amidar and Alien are both famous games on the Atari2600.
%In Amidar, the agent needs to avoid enemies and walk through all edges.
%In Alien, the agent needs to avoid enemies and collect all the yellow balls, similar to in the game Pacman.
%However, in Alien, the agent can use a weapon to attack enemies.
%This means that the agent needs to learn not only how to avoid but also how to attack enemies.
%Furthermore, due to enemy movements, the agent has to change the sub-policy rapidly and accurately.
We also validate TEMPLE on Atari games.
In this experiment, we compare TEMPLE with the basic PPO algorithm, LSTM policy, A2OC, FuN, and MLSH.
For all HRL approaches, we set the internal action dimension/number of sub-policies to 4.
The sequence lengths of TEMPLE are set to 32.
We run experiments on Atari2600 games using the OpenAI-Gym toolkit \cite{openaigym16}.
%To make the learning process faster, we use 128-bit ram data as the state in our experiments instead of the raw pixel inputs.% Running experiments on more Atari games with raw pixel inputs is a direction for future work.
We run each method for 5 million frames and the rewards are clipped to [$-1$, +1], as in other related works using Atari experiments.

%\begin{figure}[ht!]
%	\centering
%	\includegraphics[width=3in]{pic/comp_atari_legend.pdf}\\
%	\subfigure[On Amidar]{
%		\includegraphics[width=0.42\textwidth, height=1.6in]{pic/comp_amidar.pdf}
%	}
%	\subfigure[On Alien]{
%		\includegraphics[width=0.42\textwidth, height=1.6in]{pic/comp_alien.pdf}
%	}\vspace{-1em}
%	\caption{Experiments on Atari games}
%	\label{fig:atari_result}%
%\end{figure}
\begin{table}[ht]
	\centering
	\resizebox{0.9\linewidth}{!}{
	\begin{tabular}{lcccc}
		\toprule
		Methods  & Alien & Amidar & Seaquest & Breakout \\
		\toprule
		PPO   & 198  & 22  & 220 & 15   \\
		LSTM   & 223  & 31  & 244 & 14   \\
		FuN   & 217  & 23  & 236 & 13   \\
		MLSH       & 235  & 37  & 310 & 15    \\
		A2OC    & 240  & 37  & 322 & 17   \\
		TEMPLE-fix  & 212  & 27  & 232 & 16    \\
	    TEMPLE       & \textbf{285}  & \textbf{42} & \textbf{365} & \textbf{19}    \\
		\bottomrule
	\end{tabular}
	}
	%\vspace{-0.2em}
	\caption{Experiments on Atari games}
	\label{tab:atari_result}
	%\vspace{-0.5em}
\end{table}

The results are shown in Table \ref{tab:atari_result}.
TEMPLE achieves superior performance to other methods on all these environments.

\section{Conclusion}
In HRL a high-level policy needs to make macro decisions only in a low frequency.
However, the exact frequency is hard to be simply determined.
Previous HRL approaches employed a fixed-time skip strategy or learned a terminal condition without taking account of the context, which requires manual adjustments and also sacrifices some decision granularity.
In this paper, we propose the \emph{temporal-adaptive hierarchical policy learning} (TEMPLE) structure to adaptively control the high-level policy decision frequency.
We further analyze the change in the internal action and temporal signal during the test phase.
The results indicate that the temporal gate can lead to an adaptive switch of the high-level policy decision, and the internal action is also interpretable. 
We test TEMPLE on different types of environments, showing the generality of TEMPLE. The proposed temporal switch may serve as a standard module in HRL approaches.

%\clearpage
\bibliographystyle{named}
\fontsize{10.5pt}{10.0pt} \selectfont
\bibliography{ijcai20}

\end{document}